\documentclass[11pt, a4paper, logo, two column]{craftjarvis}

\pdftrailerid{redacted}

\makeatletter
\renewcommand\bibentry[1]{\nocite{#1}{\frenchspacing\@nameuse{BR@r@#1\@extra@b@citeb}}}
\makeatother

\usepackage{xcolor}
\definecolor{Gray}{gray}{0.9}
\definecolor{mygreen}{rgb}{0.0, 0.5, 0.0}
\definecolor{myred}{rgb}{0.8, 0.25, 0.33}
\definecolor{myblue}{rgb}{0.19, 0.55, 0.91}
\definecolor{uclablue}{rgb}{0.15, 0.45, 0.68}
\definecolor{boxgreen}{rgb}{0.02, 0.66, 0.02}
\definecolor{boxred}{rgb}{0.66, 0.1, 0.1}
\definecolor{boxblue}{rgb}{0.01, 0.01, 0.73}
\definecolor{mygray}{gray}{0.4}
\usepackage{notation}
\usepackage{times}
\usepackage{cuted}
\usepackage{arydshln}
\usepackage{hyperref}
\usepackage{marvosym}
\usepackage{graphicx}
\usepackage{amssymb}
\usepackage{url}

\usepackage{microtype}
\usepackage{subfigure}
\usepackage{booktabs}

\usepackage{amsmath}
\usepackage{mathtools}
\usepackage{amsthm}

\usepackage{listings}
\usepackage{etoc}
\usepackage{algorithm}
\usepackage{algorithmic}
\usepackage{lipsum}
\usepackage{pifont}
\usepackage{wrapfig}
\usepackage{colortbl}
\usepackage{acronym}
\usepackage{multicol}
\usepackage{multirow}
\usepackage{makecell}
\usepackage{enumitem}
\usepackage{tabularx}
\usepackage{colortbl}
\usepackage{array}
\usepackage[skip=3pt,font=small]{caption}
\usepackage{xspace}
\usepackage{mdframed}

\usepackage[export]{adjustbox}

\newcolumntype{Y}{>{\arraybackslash}X}

\usepackage[utf8]{inputenc} 
\usepackage[T1]{fontenc}    
\usepackage{hyperref}       
\usepackage{amsfonts}       
\usepackage{nicefrac}       

\usepackage{mathtools}
\usepackage{amssymb}
\usepackage{amsthm}

\renewcommand{\paragraph}[1]{\noindent\textbf{#1.}}

\renewcommand{\paragraph}[1]{\noindent\textbf{#1.}}

\makeatletter
\DeclareRobustCommand\onedot{\futurelet\@let@token\@onedot}
\def\@onedot{\ifx\@let@token.\else.\null\fi\xspace}

\makeatother

\acrodef{llms}[LLMs]{Large Language Models}
\acrodef{mlms}[MLMs]{Multimodal Language Models}
\newcommand{\agent}{\textsc{ROCKET-1}\xspace}

\usepackage{xcolor}

\usepackage[authoryear, sort&compress, round]{natbib} 

\title{
ROCKET-1: Mastering Open-World Interaction with Visual-Temporal Context Prompting
}

\reportnumber{} 

\renewcommand{\today}{} 

\author[1]{Shaofei~Cai}
\author[1]{Zihao~Wang}
\author[1]{Kewei~Lian}
\author[1]{Zhancun~Mu}
\author[3]{Xiaojian~Ma}
\author[2]{Anji~Liu}
\author[1]{Yitao~Liang \textrm{\Letter}}

\affil[1]{PKU}
\affil[2]{UCLA}
\affil[3]{BIGAI}
\affil[ \hspace{-0.73ex}]{All authors are affiliated with Team CraftJarvis} 

\begin{abstract}
Vision-language models (VLMs) have excelled in multimodal tasks, but adapting them to embodied decision-making in open-world environments presents challenges. 
One critical issue is bridging the gap between discrete entities in low-level observations and the abstract concepts required for effective planning. 
A common solution is building hierarchical agents, where VLMs serve as high-level reasoners that break down tasks into executable sub-tasks, typically specified using language. 
However, language suffers from the inability to communicate detailed spatial information. 
We propose visual-temporal context prompting, a novel communication protocol between VLMs and policy models. This protocol leverages object segmentation from past observations to guide policy-environment interactions. Using this approach, we train \agent, a low-level policy that predicts actions based on concatenated visual observations and segmentation masks, supported by real-time object tracking from SAM-2. Our method unlocks the potential of VLMs, enabling them to tackle complex tasks that demand spatial reasoning. 
Experiments in Minecraft show that our approach enables agents to achieve previously unattainable tasks, with a $\mathbf{76}\%$ absolute improvement in open-world interaction performance. Codes and demos are now available on the project page: \url{https://craftjarvis.github.io/ROCKET-1}. 
\end{abstract}

\begin{document}

\correspondingauthor{Yitao~Liang\\
Shaofei Cai <caishaofei@stu.pku.edu.cn>, Zihao Wang <zhwang@stu.pku.edu.cn>, Kewei Lian <lkwkwl@stu.pku.edu.cn>, Zhancun Mu <muzhancun@stu.pku.edu.cn>, Xiaojian~Ma <xiaojian.ma@ucla.edu>, Anji Liu <liuanji@cs.ucla.edu>, Yitao Liang <yitaol@pku.edu.cn>
}

\twocolumn[{%
\renewcommand\twocolumn[1][]{#1}%
\maketitle
\centering
\includegraphics[width=0.99\linewidth]{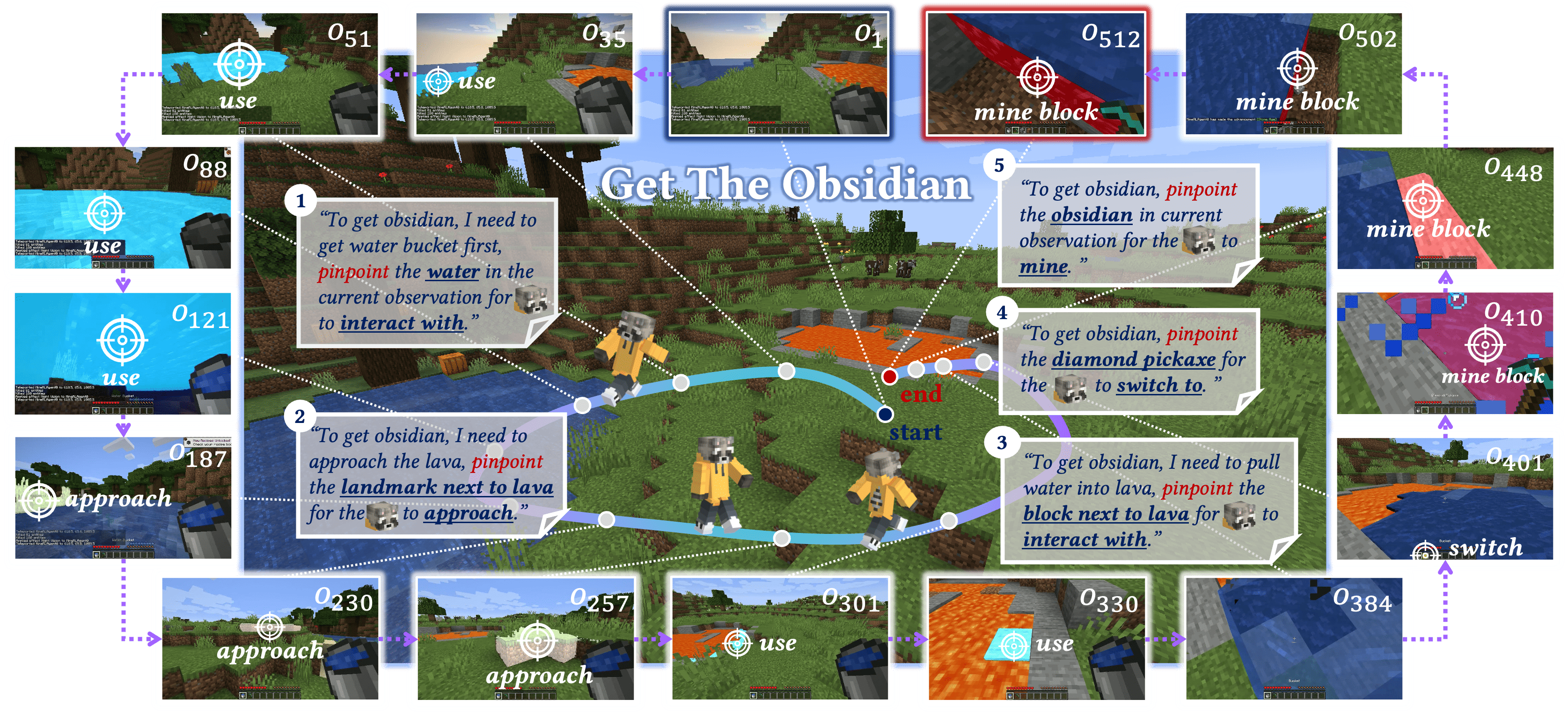}
\captionof{figure}{
Our pipeline solves creative tasks, such as \textit{get the obsidian} in the original Minecraft version, \emph{using the action space identical to human players (mouse and keyboard)}. 
We present a novel instruction interface, \emph{visual-temporal context prompting}, under which we learn a spatial-sensitive policy, \agent. VLMs identify regions of interest within each observation and guide \agent interactions. Different colors in the segmentation represent distinct interaction types, for example,  \textcolor[HTML]{90E0EF}{\rule{0.25cm}{0.25cm}} - use, \textcolor[HTML]{E5E5E5}{\rule{0.25cm}{0.25cm}} - approach, \textcolor[HTML]{FCBF49}{\rule{0.25cm}{0.25cm}} - switch, \textcolor[HTML]{FFAFCC}{\rule{0.25cm}{0.25cm}} - mine block. 
\vspace{1em}
}
\label{fig:teaser}
}]

\begin{figure*}[t]
\begin{center}
\includegraphics[width=0.95\linewidth]{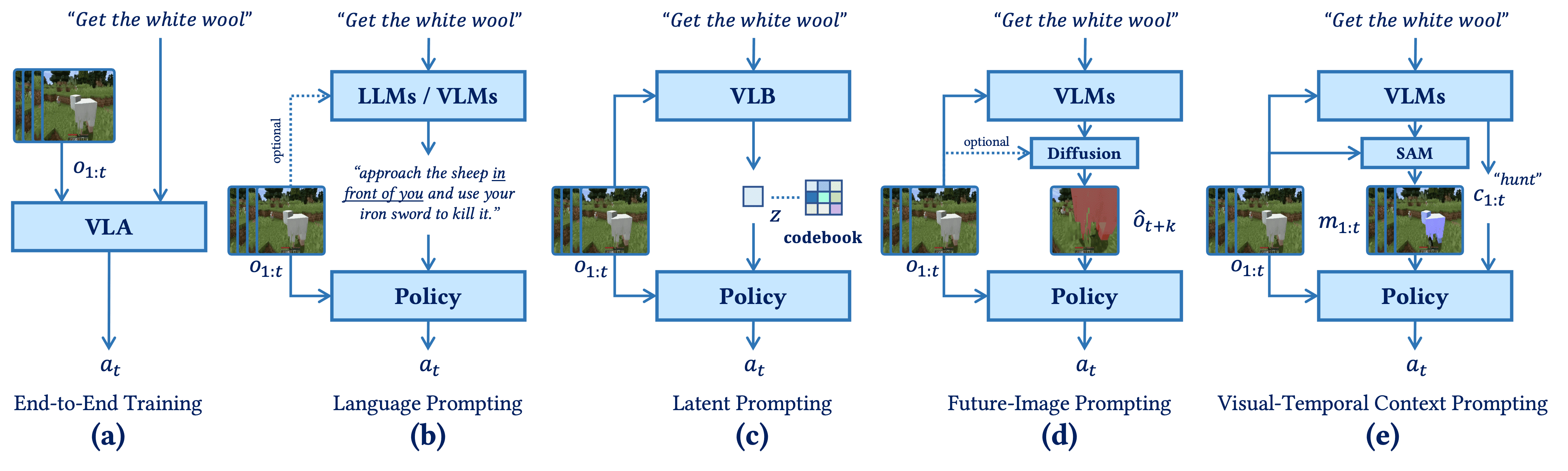}
\end{center}
\vspace{-2.0 mm}
\caption{
\textbf{Different pipelines in solving embodied decision-making tasks.} 
\textbf{(a)} End-to-end pipeline modeling token sequences of language, observations, and actions. 
\textbf{(b)} Language prompting: VLMs decompose instructions for language-conditioned policy execution. 
\textbf{(c)} Latent prompting: maps discrete behavior tokens to low-level actions. 
\textbf{(d)} Future-image prompting: fine-tunes VLMs and diffusion models for image-conditioned control. 
\textbf{(e)} Visual-temporal prompting: VLMs generate segmentations and interaction cues to guide \agent. 
}
\label{fig:comparison}
\end{figure*}

\section{Introduction}
\label{sec:intro}

Pre-trained foundation vision-language models (VLMs) \citep{gemini, gpt4} have shown impressive performance in reasoning, visual question answering, and task planning \citep{rt-2,palm-e,deps,agentsurvey}, primarily due to training on internet-scale multimodal data. Recently, there has been growing interest in transferring these capabilities to embodied decision-making in open-world environments. Existing approaches can be broadly categorized into (i) end-to-end and (ii) hierarchical approaches. End-to-end approaches, such as RT-2 \citep{rt-2}, Octo \citep{octo}, LEO \citep{leo}, and OpenVLA \citep{openvlm}, aim to enable VLMs to interact with environments by collecting robot manipulation trajectory data annotated with text. This data is then tokenized to fine-tune VLMs into vision-language-action models (VLAs) in an end-to-end manner, as illustrated in Figure \ref{fig:comparison}(a). However, collecting such annotated trajectory data is difficult to scale. Moreover, introducing the action modality risks compromising the foundational abilities of VLMs. 

Hierarchical agent architectures typically consist of a high-level reasoner and a low-level policy, which can be trained independently. In this architecture, the ``communication protocol'' between components defines the capability limits of the agent. Alternative approaches \citep{deps, voyager, palm-e} leverage VLMs’ reasoning abilities to zero-shot decompose tasks into language-based sub-tasks, with a separate language-conditioned policy executing them in the environment, refer to Figure \ref{fig:comparison}(b). However, language instructions often fail to effectively convey spatial information, limiting the tasks agents can solve. For example, when multiple homonymous objects appear in an observation image, distinguishing a specific one using language alone may require extensive spatial descriptors, increasing data collection complexity and learning difficulty for the language-conditioned policy. 
To address this issue, approaches like STEVE-1 \citep{steve1}, GROOT-1 \citep{groot1}, and MineDreamer \citep{minedreamer} propose using a purely vision-based interface to convey task information to the low-level policy. MineDreamer, in particular, uses hindsight relabeling to train an image-conditioned policy \citep{steve1} for interaction, while jointly fine-tuning VLMs and diffusion models to generate goal images that guide the policy, shown in Figure \ref{fig:comparison}(d). Although replacing language with imagined images as the task interface simplifies data collection and policy learning, predicting future observations requires building a world model, which still faces challenges such as hallucinations, temporal inconsistencies, and limited temporal scope. 

In human task execution, such as object grasping, people do not pre-imagine holding an object but maintain focus on the target object while approaching its affordance. When the object is obscured, humans rely on memory to recall its location and connect past and present visual scenes. This use of visual-temporal context enables humans to solve tasks effectively in novel environments. 
Building on this idea, we propose a novel communication protocol called \textbf{visual-temporal context prompting}, as shown in Figure \ref{fig:comparison}(e). This allows users/reasoners to apply object segmentation to highlight regions of interest in past visual observations and convey interaction-type cues via a set of skill primitives. Based on this, we learn \textbf{\agent}, a low-level policy that uses visual observations and reasoner-provided segmentations as task prompts to predict actions causally. Specifically, a transformer \citep{transformerxl} models dependencies between observations, essential for representing tasks in partially observable environments. As a bonus feature, \agent can enhance its object-tracking capabilities during inference by integrating the state-of-the-art video segmentation model, SAM-2 \citep{sam2}, in a plug-and-play fashion. Additionally, we propose a \textbf{backward trajectory relabeling} method, which efficiently generates segmentation annotations in reverse temporal order using SAM-2, facilitating the creation of training datasets for \agent. 
Finally, we develop a hierarchical agent architecture leveraging visual-temporal context prompting, which perfectly inherits the vision-language reasoning capabilities of foundational VLMs. Experiments in Minecraft demonstrate that our pipeline enables agents to complete tasks previously unattainable by other methods, while the hierarchical architecture effectively solves long-horizon tasks. 

Our main contributions are threefold:
\textbf{(1)} We present \textbf{visual-temporal context prompting}, a novel protocol that effectively communicates spatial and interaction cues in hierarchical agent architecture. 
\textbf{(2)} We learn \textbf{\agent}, the first segmentation-conditioned policy in Minecraft, capable of interacting with nearly all the objects. 
\textbf{(3)} We develop \textbf{backward trajectory relabeling} method that can automatically detect and segment desired objects in collected trajectories with pre-trained SAMs for training \agent.

\section{Preliminaries}
\label{sec:preliminaries}

\paragraph{Offline Reinforcement Learning}
We model the open-world interaction problem as a Markov Decision Process (MDP) $\left< \mathcal{O}, \mathcal{A}, \mathcal{P}, \mathcal{C}, \mathcal{M}, \mathcal{R} \right>$, where $\mathcal{O}$ and $\mathcal{A}$ represent the observation and action spaces, $\mathcal{P}: \mathcal{O} \times \mathcal{A} \times \mathcal{O} \rightarrow \mathbb{R}^{+}$ describes the environment dynamics, $\mathcal{C}$ is the set of interaction types, and $\mathcal{M}$ is the segmentation mask space. The binary reward function $\mathcal{R}: \mathcal{O} \times \mathcal{A} \times \mathcal{C} \times \mathcal{M} \rightarrow \{0, 1\}$ determines whether the policy has completed the specified interaction with the object indicated by the segmentation mask at each time step. The objective of reinforcement learning is to learn a policy that maximizes the expected cumulative reward, $\mathbb{E} \left[ \sum_{t=1}^{T} r_t \right]$, where $r_t$ is the reward at time step $t$. 
Our proposed \emph{backward trajectory relabeling} method ensures that each trajectory attains a positive reward based on current object segmentations. This allows us to discard the rewards and learn a conditioned policy $\pi(a | o, c, m)$ directly using behavior cloning. In the offline setting, agents do not interact with the environment but rely on a fixed, limited dataset of trajectories. This setting is harder as it removes the ability to explore the environment and gather additional feedback. 

\paragraph{Vision Language Models} 
Vision-Language Models (VLMs) are machine learning models capable of processing both image and language modalities. Recent advances in generative pretraining have led to the emergence of conversational models like Gemini \citep{gemini}, GPT-4o \citep{gpt4}, and Molmo \citep{molmo}, which are trained on large-scale multimodal data and can reason and generate human-like responses based on text and images. Models such as Palm-E \citep{palm-e} have demonstrated strong abilities in embodied question-answering and task planning. However, standalone VLMs cannot often interact directly with environments. Some approaches use VLMs to generate language instructions for driving low-level controllers, but these methods struggle with expressing spatial information. This work focuses on releasing VLMs’ spatial understanding in embodied decision-making scenarios. 
Molmo can accurately identify correlated objects in images using a list of $(x, y)$ coordinates, as demonstrated in \url{https://molmo.allenai.org}. 

\begin{figure*}[ht]
\begin{center}
\includegraphics[width=0.99\linewidth]{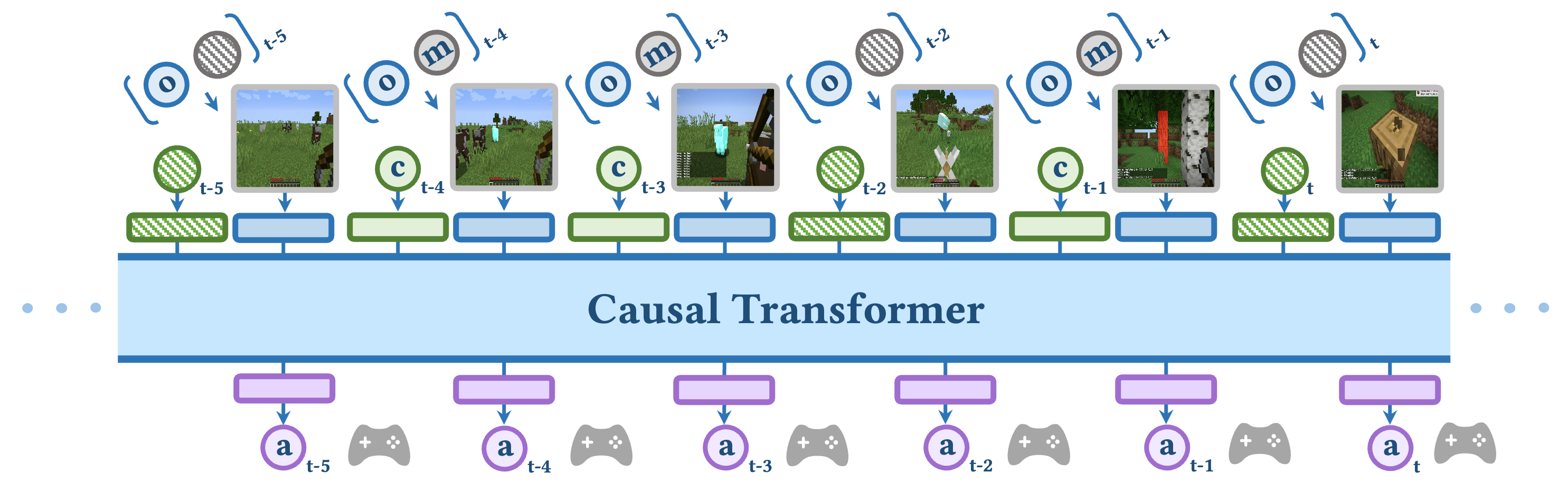}
\end{center}
\vspace{-2mm}
\caption{
\textbf{\agent architecture.} \agent processes observations ($o$), object segmentations ($m$), and interaction types ($c$) to predict actions ($a$) using a causal transformer. Observations and segmentations are concatenated and passed through a visual backbone for deep fusion. Interaction types and segmentations are randomly dropped with a pre-defiened probability during training. 
}
\label{fig:pipeline}
\end{figure*}

\paragraph{Segment Anything Models} 
The Segment Anything Model (SAM, \cite{sam}), introduced by Meta, is a segmentation model capable of interactively segmenting objects based on point or bounding box prompts, or segmenting all objects in an image at once. It demonstrates impressive zero-shot generalization in both real-world and video game environments. Recently, Meta introduced SAM-2 \citep{sam2}, extending segmentation to the temporal domain. With SAM-2, users can prompt object segmentation with points or bounding boxes on a single video frame, and the model will track the object forward or backward in time, refer to \url{https://ai.meta.com/sam2}. Remarkably, SAM-2 continues tracking even if the object disappears and reappears, making it well-suited for partially observable open-world environments. In addition, we find the SAM models can be equipped with a text prompt module, enabling them to ground text-based concepts in visual images, as seen in grounded SAM \citep{liu2023grounding}.

\section{Methods}
\label{sec:method}

\paragraph{Overview} Our work focuses on addressing complex interactive tasks in open-world environments like Minecraft. We leverage VLMs’ visual-language reasoning capabilities to decompose tasks into multiple steps and determine object interactions based on environmental observations. For example, the \textit{``build nether portal''} task requires a sequence of block placements at specific locations. A controller is also needed to map these steps into low-level actions. To convey spatial information accurately, we propose a visual-temporal context prompting protocol and a low-level policy, \agent. Pretrained VLMs process a sequence of frames $o_{1:t}$  and a language-based task description to generate object segmentations $m_{1:t}$  and interaction types $c_{1:t}$, representing the interaction steps. The learned \agent $\pi(a_t | o_{1:t}, m_{1:t}, c_{1:t})$ interprets these outputs to interact with the environment in real-time. In this section, we outline \agent’s architecture and training methods, the dataset collection process, and a pipeline integrating \agent with state-of-the-art VLMs. 

\begin{figure*}[t]
\begin{center}
\includegraphics[width=0.95\linewidth]{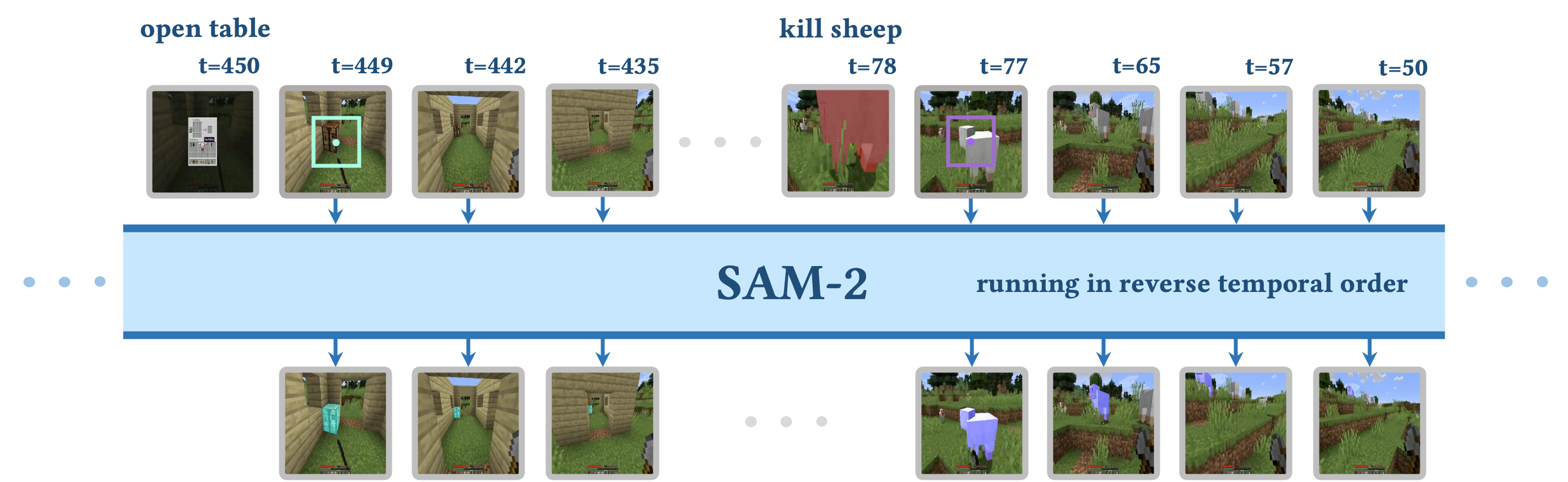}
\end{center}
\caption{
\textbf{Trajectory relabeling pipeline in Minecraft.} A bounding box and point selection are applied to the image center in the frame preceding the interaction event to identify the interacting object. SAM-2 is then run in reverse temporal order for a specified duration.
} 
\label{fig:datapipe}
\end{figure*}

\paragraph{\agent Architecture}
To train \agent, we prepare interaction trajectory data in the format:
$\tau = (o_{1:T}, a_{1:T}, m_{1:T}, c_{1:T})$,
where $o_t \in \mathbb{R}^{3 \times H \times W}$ is the visual observation at time $t$, $m_t \in \{0, 1\}^{1 \times H \times W}$ is a binary mask highlighting the object in $o_t$ for future interaction, $c_t \in \mathbb{N}$ denotes the interaction type, and $a_t$ is the action. If both $m_t$ and $c_t$ are zeros, no region is highlighted at $o_t$. 
As shown in Figure \ref{fig:pipeline}, \agent is formalized as a conditioned policy, $\pi(a_t | o_{1:t}, m_{1:t}, c_{1:t})$, which takes a sequence of observations and object-segmented interaction regions to causally predict actions. To effectively encode spatial information, inspired by \cite{controlnet}, we concatenate the observation and object segmentation pixel-wise into a 4-channel image, which is processed by a visual backbone for deep fusion, followed by an self-attention pooling layer: 
\begin{align}
h_t &\leftarrow \texttt{Backbone}([o_t, m_t]), \\
x_t &\leftarrow \texttt{AttentionPooling}(h_t). 
\end{align}
We extend the input channels of the first convolution in the pre-trained visual backbone from 3 to 4, initializing the new parameters to $0$s to minimize the gap in early training. A TransformerXL \citep{transformerxl, vpt} module is then used to model temporal dependencies between observations and incorporate interaction type information to predict the next action $\hat{a}_t$: 
\begin{equation}
    \hat{a}_t \leftarrow \texttt{TransformerXL}(c_1, x_1, \cdots, c_t, x_t). 
\end{equation}
We delay the integration of interaction type information $c_t$ until after fusing $m_t$ and $o_t$, enabling the backbone to share knowledge across interaction types and mitigating data imbalance. Behavior cloning loss is used for optimization. However, this approach risks making $a_t$ overly dependent on $m_t$ and $c_t$, reducing the model's temporal reasoning capability. To address this, we propose randomly dropping segmentations with a certain probability, forcing the model to infer user intent from past inputs (visual-temporal context). 
The final optimization objective is: 
\begin{equation}
    \mathcal{L} = -\sum_{t=1}^{|\tau|} \log \pi(a_t | o_{1:t}, m_{1:t} \odot w_{1:t}, c_{1:t} \odot w_{1:t}), 
\end{equation}
where $w_t \sim \text{Bernoulli}(1-p)$ represents a mask, with $p$ denoting the dropping probability, $\odot$ denotes the product operation over time dimension. 

\paragraph{Backward Trajectory Relabeling} We seek to build a dataset for training \agent. The collected trajectory data $\tau$ typically \emph{contains only observations $o_{1:T}$ and actions $a_{1:T}$.} To generate object segmentations and interaction types for each frame, we propose \emph{a novel hindsight relabeling technique} \citep{her} combined with an object tracking model \citep{sam2} for automatic data labeling. We first abstract a set of interactions $\mathcal{C}$ and identify frames where interaction events occur, detected using a pre-trained vision-language model, such as \cite{gpt4}. Then, we traverse the trajectory in reverse order, segmenting interacting objects in frame $t$ via an open-vocabulary grounding model, such as \citep{liu2023grounding}. Finally, SAM-2 \citep{sam2} is used to track and generate segmentations for frames $t-1, t-2, \dots, t-k$, where $k$ is the window length. 

For Minecraft, we use contractor data \citep{vpt} from OpenAI, consisting of 1.6 billion frames of human gameplay. This dataset includes meta information for each frame, recording interaction events such as \textit{kill entity}, \textit{mine block}, \textit{use item}, \textit{interact}, \textit{craft}, and \textit{switch}, eliminating the need for vision-language models to detect events. We observed that interacting objects are often centered in the previous frame, allowing the use of a fixed-position bounding box and point with the SAM-2 model for segmentation, replacing open-vocabulary grounding models. 
We also introduced an additional interaction type, \textit{navigate}.
If a player’s movement exceeds a set threshold over a period, they are considered to be approaching an object. The object they face in the segment’s final frame is marked as the target, with SAM-2 applied in reverse to identify it in earlier frames. As shown in Figure \ref{fig:datapipe}, the entire labeling process can be totally automated. 

\begin{figure*}[t]
\begin{center}
\includegraphics[width=0.95\linewidth]{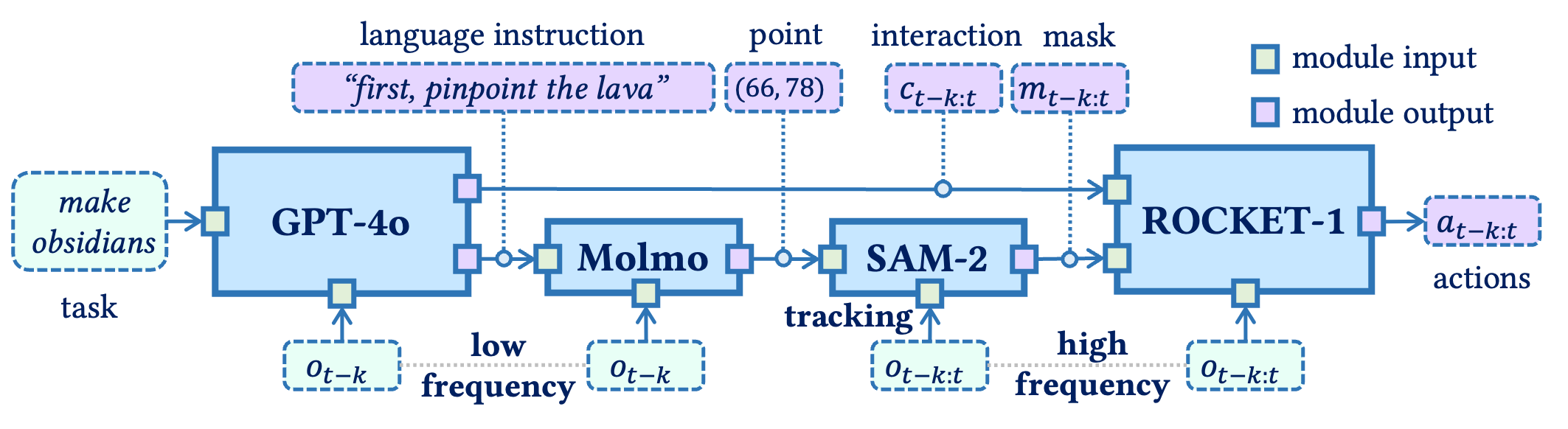}
\end{center}
\vspace{-3mm}
\caption{
\textbf{A hierarchical agent structure based on our proposed visual-temporal context prompting. } A GPT-4o model decomposes complex tasks into steps based on the current observation, while the Molmo model identifies interactive objects by outputting points. SAM-2 segments these objects based on the point prompts, and \agent uses the object masks and interaction types to make decisions. GPT-4o and Molmo run at low frequencies, while SAM-2 and \agent operate at the same frequency as the environment. 
} 
\label{fig:inference}
\end{figure*}

\paragraph{Integration with High-level Reasoner} 
Completing complex long-horizon tasks in open-world environments requires agents to have strong commonsense knowledge and do visual-language reasoning, both of which are strengths of modern VLMs. As shown in Figure \ref{fig:inference}, we design a novel hierarchical agent architecture comprising GPT-4o \citep{gpt4}, Molmo \citep{molmo}, SAM-2 \citep{sam2}, and ROCKET-1. GPT-4o decomposes tasks into object interactions based on an observation $o_{t-k}$, leveraging its extensive knowledge and reasoning abilities. Since GPT-4o cannot directly output the object masks, we use Molmo to generate $(x, y)$ coordinates for the described objects. SAM-2 then produces the object mask $m_{t-k}$ from these coordinates and efficiently tracks objects $m_{t-k+1:t}$ in subsequent observations. ROCKET-1 uses the generated masks $m_{t-k:t}$ and interaction types $c_{t-k:t}$ from GPT-4o to engage with the environment. Due to the high computational cost, GPT-4o and Molmo run at lower frequencies, while SAM-2 and ROCKET-1 operate at the env’s frequency. 

\begin{table}[t]
\small
\centering
\renewcommand{\arraystretch}{1.1}
\caption{
\textbf{Hyperparameters for training \agent.} 
} \label{table:implementation}
\begin{adjustbox}{width=\linewidth}
\begin{tabular}{@{}lc@{}}
\toprule
\textbf{Hyperparameter}  & \textbf{Value}   \\ \midrule
Input Image Size         & $224 \times 224$ \\
Visual Backbone          & EfficientNet-B0 (4 channels)  \\
Policy Transformer       & TransformerXL \\
Number of Policy Blocks  & $4$         \\
Hidden Dimension         & $1024$      \\
Trajectory Chunk size    & $128$       \\
Dropout Rate $p$         & $0.75$      \\
Optimizer                & AdamW       \\
Learning Rate            & $0.00004$   \\
\bottomrule
\end{tabular}
\end{adjustbox}
\end{table}

\section{Results and Analysis}
\label{sec:results}

First, we provide a detailed overview of the experimental setup, including the benchmarks, baselines, and implementation details. We then explore \agent’s performance on basic open-world interactions and long-horizon tasks. Finally, we conduct comprehensive ablation studies to validate the rationale behind our design choices. 

\begin{figure*}[t]
\begin{center}
\includegraphics[width=0.95\linewidth]{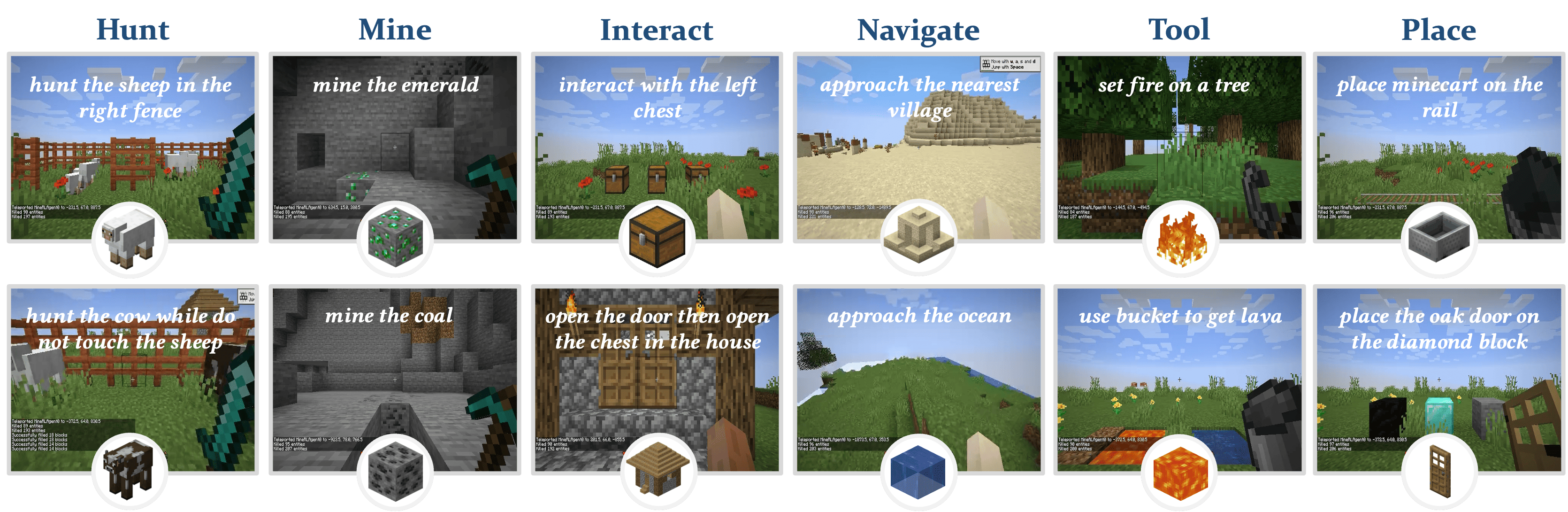}
\end{center}
\vspace{-2 mm}
\caption{
\textbf{A benchmark for evaluating open-world interaction capabilities of agents. } The benchmark contains six interaction types in Minecraft, totaling 12 tasks. Unlike previous benchmarks, these tasks emphasize interacting with objects at specific spatial locations. For example, in \textit{``hunt the sheep in the right fence,''} the task fails if the agent kills the sheep on the left side. Some tasks, such as \textit{``place the oak door on the diamond block,''} never appear in the training set. It is also designed to evaluate zero-shot generalization capabilities. 
} 
\label{fig:benchmark}
\end{figure*}

\begin{table*}[ht]
\centering
\renewcommand{\arraystretch}{1.1}
\small
\caption{
\textbf{Results on the Minecraft Interaction benchmark. } Each task is tested 32 times, and the average success rate is reported as the final result. ``Human'' indicates instructions provided by a human. 
} \label{tab:results}
\begin{adjustbox}{width=\linewidth}
\begin{tabular}{@{}lcccccccccccccc@{}}
\toprule
\multirow{2}{*}{\textbf{Method}} & \multirow{2}{*}{\textbf{Prompt}} & \multicolumn{2}{c}{\textbf{Hunt}} & \multicolumn{2}{c}{\textbf{Mine}} & \multicolumn{2}{c}{\textbf{Interact}} & \multicolumn{2}{c}{\textbf{Navigate}} & \multicolumn{2}{c}{\textbf{Tool}} & \multicolumn{2}{c}{\textbf{Place}} & \multirow{2}{*}{\textbf{Avg}} \\ \cmidrule(lr){3-14}
                        &                         &  \includegraphics[scale=0.5,valign=c]{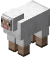}           &  \includegraphics[scale=0.45,valign=c]{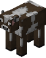}          &  \includegraphics[scale=0.45,valign=c]{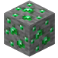}           & \includegraphics[scale=0.45,valign=c]{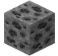}           & \includegraphics[scale=0.45,valign=c]{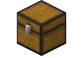}      &  \includegraphics[scale=0.45,valign=c]{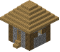}      & \includegraphics[scale=0.45,valign=c]{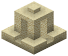}              & \includegraphics[scale=0.45,valign=c]{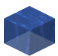}             & \includegraphics[scale=0.45,valign=c]{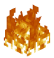}            & \includegraphics[scale=0.45,valign=c]{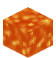}           & \includegraphics[scale=0.45,valign=c]{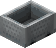}            & \includegraphics[scale=0.45,valign=c]{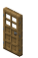}            &                          \\ \midrule
VPT-bc                & N/A                     &   0.13          &  0.16          &  0.00           &  0.13          &  0.03             &  0.31            &  0.00             &  0.09            &  0.00           &   0.00         &  0.00           &  0.00     &  0.07                       \\
STEVE-1  & Human & 0.00  & 0.06  & 0.00  & 0.69 & 0.00  & 0.03  & 0.00  & 0.31 & 0.91 & 0.06  & 0.16 & 0.00  &  0.19 \\
GROOT-1  & Human & 0.09  & 0.22 & 0.00  & 0.06  & 0.03  & 0.06  & 0.00  & 0.03  & 0.47 & 0.13 & 0.03  & 0.00  &  0.09 \\  
 ROCKET-1   & Molmo  &  $\mathbf{0.91}$ &  $\mathbf{0.84}$ &  $\mathbf{0.78}$ & $\mathbf{0.75}$ & $\mathbf{0.81}$ & $\mathbf{0.50}$ & $\mathbf{0.78}$ &  $\mathbf{0.97}$ & $\mathbf{0.94}$ &  $\mathbf{0.91}$ & $\mathbf{0.72}$ & $\mathbf{0.91}$ & $\mathbf{0.82}$ \\
 ROCKET-1   & Human  & $\mathbf{0.94}$ & $\mathbf{0.91}$ & $\mathbf{0.91}$ & $\mathbf{0.94}$ & $\mathbf{0.94}$ & $\mathbf{0.91}$ & $\mathbf{0.97}$ & $\mathbf{0.97}$ & $\mathbf{0.97}$ & $\mathbf{0.97}$ &  $\mathbf{0.94}$ & $\mathbf{0.97}$ & $\mathbf{0.95}$ \\
 \bottomrule
\end{tabular}
\end{adjustbox}
\end{table*}

\subsection{Experimental Setup}

\paragraph{Implementation Details} 
Briefly, we present \agent’s model architecture, hyperparameters, and optimizer configurations in Table \ref{table:implementation}. During training, each complete trajectory is divided into 128-length segments to reduce memory requirements. During inference, \agent can access up to 128 frames of past observations. Most training parameters follow the settings from prior works such as \cite{groot1, groot2, vpt}. 

\paragraph{Environment and Benchmark} We use the unmodified Minecraft 1.16.5 \citep{minerl,mcu} as our testing environment, which accepts mouse and keyboard inputs as the action space and outputs a $640\times360$ RGB image as the observation. To comprehensively evaluate the agent’s interaction capabilities, as shown in Figure \ref{fig:benchmark}, we introduce the \textbf{Minecraft Interaction Benchmark}, consisting of six categories and a total of 12 tasks, including \textit{Hunt}, \textit{Mine}, \textit{Interact}, \textit{Navigate}, \textit{Tool}, and \textit{Place}. This benchmark emphasizes object interaction and spatial localization skills. For example, in the \textit{``hunt the sheep in the right fence''} task, success requires the agent to kill sheep within the right fence, while doing so in the left fence results in failure. In the \textit{``place the oak door on the diamond block''} task, success is achieved only if the oak door is adjacent to the diamond block on at least one side. 

\paragraph{Baselines} We compare our methods with the following baselines: (1) VPT \citep{vpt}: A foundational model pre-trained on large-scale YouTube data, with three variants—VPT (fd), VPT (bc), and VPT (rl)—representing the vanilla foundational model, behavior-cloning finetuned model, and RL-finetuned model, respectively. In this study, we utilize the VPT (bc) variant. 
(2) STEVE-1 \citep{steve1}: An instruction-following agent finetuned from VPT, capable of solving various short-horizon tasks. We select the text-conditioned version of STEVE-1 for comparison. 
(3) GROOT-1 \citep{groot1}: A reference-video conditioned policy designed to perform open-ended tasks, trained on 2,000 hours of long-form videos using latent variable models. 

\begin{figure*}[t]
\begin{center}
\includegraphics[width=1.0\linewidth]{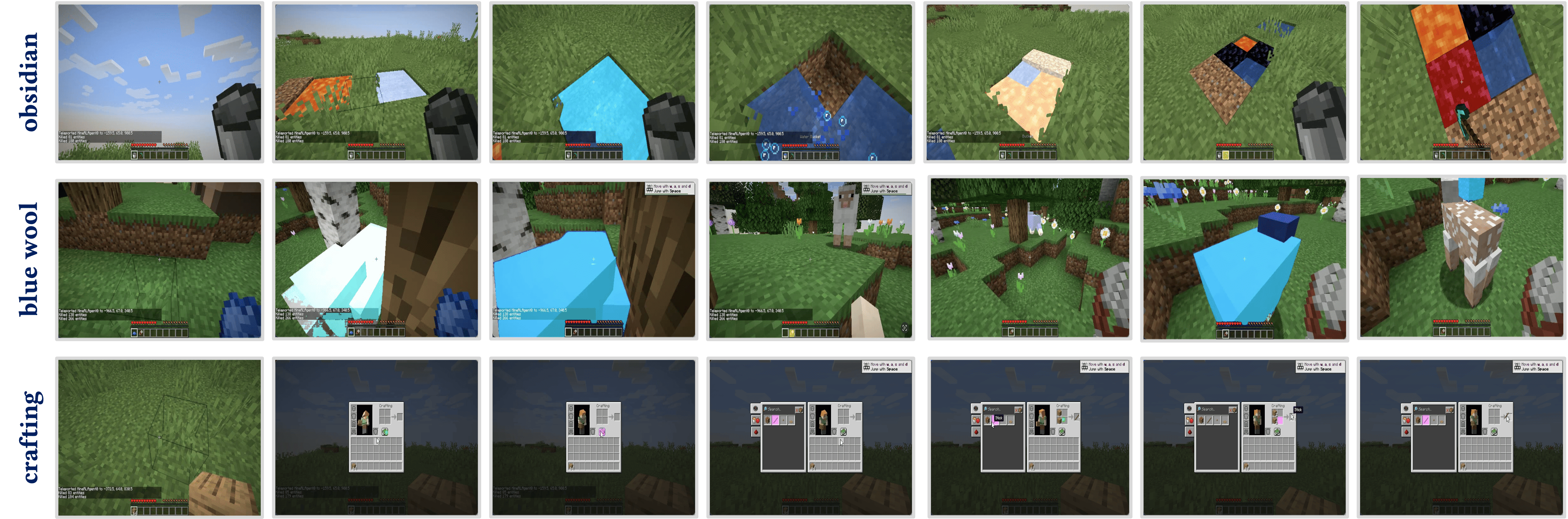}
\end{center}
\caption{
\textbf{Screenshots of our hierarchical agent when completing long-horizon tasks. } 
} 
\label{fig:demo}
\end{figure*}

\begin{table*}[]
\centering
\renewcommand{\arraystretch}{1.1}
\footnotesize
\caption{
\textbf{Comparison of hierarchical architectures with different communication protocols. } All seven tasks require complex reasoning capabilities. The diamond task was run 100 times, while other tasks were run 20 times, with average success rates reported. 
} \label{tab:long-horizon}
\begin{adjustbox}{width=\linewidth}
\begin{tabular}{@{}lccccccccc@{}}
\toprule
\textbf{Method} & \textbf{Communication Protocol} & \textbf{Policy} & \includegraphics[scale=0.20,valign=c]{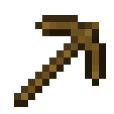} & \includegraphics[scale=0.20,valign=c]{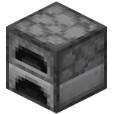} & \includegraphics[scale=0.20,valign=c]{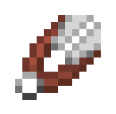} & \includegraphics[scale=0.20,valign=c]{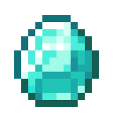} & \includegraphics[scale=0.20,valign=c]{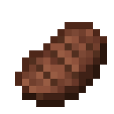} & \includegraphics[scale=0.20,valign=c]{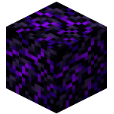} & \includegraphics[scale=0.20,valign=c]{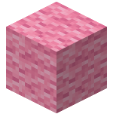} \\ \midrule
DEPS        & language                & STEVE-1  &   $0.95$              &  $0.75$       &   $0.15$     &  $0.02$            &   $0.15$        &    $0.00$      &  $0.00$         \\
MineDreamer$^*$ & future image            & STEVE-1  &  $0.95$               &   -    &   -     &    -          &   $0.00$        &  $0.00$        &  $0.00$         \\
OmniJarvis  & latent code             & GROOT-1  &  $0.95$              &  $0.90$       &  $0.20$      &   $0.08$           &    $0.40$       &  $0.00$        &   $0.00$   \\ 
\textbf{Ours}        & visual-temporal context & ROCKET-1 &  $\mathbf{1.00}$               &   $\mathbf{1.00}$      &  $\mathbf{0.45}$      &  $\mathbf{0.25}$            &  $\mathbf{0.75}$         &   $\mathbf{0.50}$       &  $\mathbf{0.70}$        \\ \bottomrule
\end{tabular}
\end{adjustbox}
\end{table*}

\subsection{ROCKET-1 Masters Minecraft Interactions }
We evaluated \agent on the Minecraft Interaction Benchmark, with results as illustrated in Table \ref{tab:results}. Since \agent operates as a low-level policy, it requires a high-level reasoner to provide prompts within a visual-temporal context, driving \agent’s interactions with the environment. We tested two reasoners: (1) A skilled Minecraft human player, who can provide prompts to \agent at any interaction moment, serving as an oracle reasoner that demonstrates the upper bound of \agent’s capabilities. (2) A Molmo 72B model \citep{molmo}, where a predefined Molmo prompt is set for each task to periodically select points in the observation as prompts, which are then processed into object segmentations by the SAM-2 model \citep{sam2}. Between Molmo’s invocations, SAM-2’s tracking capabilities offer object segmentations to guide \agent. For all baselines, humans provide prompts. We found that \agent + Molmo consistently outperformed all baselines, notably achieving a $91\%$ success rate in the \textit{``place oak door on the diamond block''} task that no baseline can solve. 

\subsection{ROCKET-1 Supports Long-Horizon Tasks}
We compared hierarchical agent architectures based on different communication protocols: (1) language-based approaches, exemplified by DEPS \citep{deps}; (2) future-image-based methods, represented by MineDreamer \citep{minedreamer}; (3) latent-code-based methods, as in OmniJarvis \citep{omnijarvis}; and (4) our proposed approach based on visual-temporal context, as illustrated in the Figure \ref{fig:inference}. 
For MineDreamer, we used the planner provided by DEPS and MineDreamer as the controller to complete the long-horizon experiment.
We evaluated these methods on seven tasks, each requiring long-horizon planning: obtaining a wooden pickaxe (3.6k), furnace (6k), shears (12k), diamond (24k), steak (6k), obsidian (24k), and pink wool (6k), where the numbers in parentheses represent the time limit. In the first five tasks, the agent starts from scratch, while for the obsidian task, we provide an empty bucket and a diamond pickaxe in advance, and for the pink wool task, we provide shears. \emph{Taking the obsidian task as an example, the player must first locate a nearby water source, fill the bucket, find a nearby lava pool, pour the water to form obsidian, and finally switch to the diamond pickaxe to mine the obsidian.} Our approach significantly improved success rates on the first five tasks, particularly achieving a $35\%$ increase in the steak task. For the last two tasks, all previous baseline methods failed, while our approach achieved a $70\%$ success rate on the wool dyeing task. Figure \ref{fig:demo} presents screenshots. 

\subsection{What Matters for Learning \agent?}
We conduct ablation studies on individual tasks of Minecraft Interaction benchmark: ``Hunt right sheep (\includegraphics[scale=0.5,valign=c]{figures/icons/hunt_sheep.png})''  and ``Mine emerald (\includegraphics[scale=0.5,valign=c]{figures/icons/mine_emerald.png})''. 

\paragraph{Condition Fusion Methods} We modified the visual backbone’s input layer from 3 to 4 channels, allowing \agent to integrate object segmentation information. For fusing interaction-type information, we explored two approaches: (1) keeping the object segmentation channel binary and encoding interaction types via an embedding layer for fusion in TransformerXL, and (2) directly encoding interaction types into the object segmentation for fusion within the visual backbone. As shown in Table \ref{tab:ablation-fusion}, the first approach significantly outperformed the second, as it allows the visual backbone to share knowledge across different interaction types and focus on recognizing objects of interest without being affected by imbalanced interaction-type distributions. 

\begin{table}[]
\centering
\caption{
\textbf{Comparison of different condition fusion methods.} 
} \label{tab:ablation-fusion}
\renewcommand{\arraystretch}{1.1}
\footnotesize
\begin{adjustbox}{width=\linewidth}
\begin{tabular}{@{}lcc@{}}
\toprule
\textbf{Fusion Positions}   & \textbf{Hunt} (\includegraphics[scale=0.4,valign=c]{figures/icons/hunt_sheep.png})$\uparrow$ & \textbf{Mine} (\includegraphics[scale=0.4,valign=c]{figures/icons/mine_emerald.png}) $\uparrow$ \\ \midrule
Fusion in transformer layer & $\mathbf{0.91}$ & $\mathbf{0.78}$ \\
Fusion in visual backbone   & $0.72$ & $0.69$ \\
\bottomrule
\end{tabular}
\end{adjustbox}
\end{table}

\begin{table}[]
\centering
\caption{
\textbf{Comparison between different SAM-2 variants.} We studied the impact of SAM-2 models of different sizes on the agent's object-tracking capability (metric: success rate) and inference speed (metric: frames per second, FPS). ``\#Pmt'' indicates the number of frames between prompts generated by Molmo. 
} \label{tab:ablation-sam}
\renewcommand{\arraystretch}{1.1}
\small
\begin{adjustbox}{width=\linewidth}
\begin{tabular}{lcccc}
\toprule
\textbf{Variants}              & \textbf{\#Pmt}     & \textbf{FPS} $\uparrow$ & \includegraphics[scale=0.4,valign=c]{figures/icons/hunt_sheep.png} $\uparrow$ & \includegraphics[scale=0.4,valign=c]{figures/icons/mine_emerald.png} $\uparrow$ \\ \midrule
baseline (w/o sam2)        &  $3$ & $0.9$ & $0.84$ & $\mathbf{0.82}$      \\
baseline (w/o sam2)        & $30$ & $\mathbf{9.2}$ & $0.00$  & $0.03$  \\
+sam2\_tiny                & $30$ & $5.4$ & $0.84$ & $0.69$ \\
+sam2\_small               & $30$ & $5.1$ & $0.88$ & $0.50$ \\
+sam2\_base\_plus          & $30$ & $3.0$ & $0.88$ & $0.63$ \\
+sam2\_large               & $30$ & $2.4$ & $\mathbf{0.91}$ & $0.78$ \\ \bottomrule
\end{tabular}
\end{adjustbox}
\end{table}

\paragraph{SAM-2 Models} The SAM-2 model acts as a proxy segmentation generator when the high-level reasoner fails to provide timely object segmentations. We evaluate the impact of different SAM-2 model sizes on task performance and inference speed, as shown in Table \ref{tab:ablation-sam}. Results indicate that with low-frequency prompts from the high-level reasoner (Molmo 72B) at $1.5$ (game frequency is 20), SAM-2 greatly improves task success rates. While ``sam2\_hiera\_large'' is the best, increasing the SAM-2 model size yields performance gains at the cost of higher time.


\section{Related Works}
\label{sec:related}

\paragraph{Instructions for Multi-Task Policy} 
Most current approaches \citep{rt-1, rt-2, language_table, worldmc, leo} use natural language to describe task details and collect large amounts of text-demonstration data pairs to train a language-conditioned policy for interaction with the environment. Although natural language can express a wide range of tasks, it struggles to represent spatial relationships effectively. Additionally, gathering text-annotated demonstration data is costly, limiting the scalability of these methods. Alternatives, such as \cite{steve1, zson, rt-sketch}, use images to drive goal-conditioned policies, typically learning through hindsight relabeling in a self-supervised manner. While this reduces the need for annotated data, future images are often insufficiently expressive, making it difficult to capture detailed task execution processes. Methods like \cite{groot1, bcz} propose using reference videos to describe tasks, offering strong expressiveness but suffering from ambiguity, which may lead to inconsistencies between policy interpretation and human understanding, raising safety concerns. \cite{rt-trajectory} suggests representing tasks with rough robot arm trajectories, enabling novel task completion but only in fully observable environments, limiting its applicability in open-world settings. CLIPort \citep{cliport}, which addresses pick-and-place tasks by controlling the robot’s start and end positions using heatmaps, bears some resemblance to our proposed visual-temporal context prompting method. However, CLIPort focuses solely on the pick-and-place task solutions in a fully observable environment.

\paragraph{Agents in Minecraft} 
Minecraft offers a highly open sandbox environment with complex tasks and free exploration, ideal for testing AGI’s adaptability and long-term planning abilities. Its rich interactions and dynamic environment simulate real-world challenges, making it an excellent testbed for AGI. 
One line of research focuses on low-level control policies in Minecraft. \cite{vpt} annotated a large YouTube Minecraft video dataset with actions and trained the first foundation agent in the domain using behavior cloning, but it lacks instruction-following capabilities. 
\cite{worldmc} employs a goal-sensitive backbone and horizon prediction module to enhance multi-task execution in partially observable environments, but it only solves tasks seen during training. \cite{minedojo} fine-tunes a vision-language alignment model MineCLIP using YouTube video data, and incorporates it into a reward shaping mechanism for training a multi-task agent, though task transfer still requires extensive environment interaction. \cite{steve1} uses hindsight-relabeling to learn an image-goal-conditioned policy and aligns image and text spaces via MineCLIP, but this approach is limited to short-horizon tasks. 
Another research focus integrates vision-language models for long-horizon task planning in Minecraft~\citep{plan4mc,rat,mp5,steveeye,liu2024rlgpt}. DEPS \citep{deps}, the first to apply large language models in Minecraft, uses a four-step process to decompose tasks, achieving the diamond mining challenge with minimal training. Voyager \citep{voyager} highlights LLM-based agents’ autonomous exploration and skill-learning abilities. Jarvis-1 \citep{jarvis-1} extends DEPS with multimodal memory, improving long-horizon task success rates by recalling past experiences. OmniJarvis \citep{omnijarvis} learns a behavior codebook using self-supervised methods to jointly model language, images, and actions. 
MineDreamer \citep{minedreamer} fine-tunes VLMs and a diffusion model to generate goal images for task execution, though it faces challenges with image quality and consistency.

\section{Conclusions and Limitations}
\label{sec:conclusion}
This paper presents a novel hierarchical agent architecture for open-world interaction. To address spatial communication challenges, we introduce visual-temporal context prompting to convey intent between the high-level reasoner and low-level policy. We develop \agent, an object-segmentation-conditioned policy for real-time object interaction, enhanced by SAM-2 for plug-and-play object tracking. Experiments in Minecraft show that our approach effectively leverages VLMs’ visual-language reasoning, achieving superior open-world interaction performance over baselines. 

Although \agent significantly enhances interaction capabilities in Minecraft, it cannot engage with objects that are outside its field of view or have not been previously encountered. For instance, if the reasoner instructs \agent to eliminate a sheep that it has not yet seen, the reasoner must indirectly guide \agent’s exploration by providing segmentations of other known objects. This limitation reduces \agent’s efficiency in completing simple tasks and necessitates frequent interventions from the reasoner, leading to increased computational overhead. We solve this problem in ROCKET-2 \citep{rocket2}. 
This project is implemented using \href{https://github.com/CraftJarvis/MineStudio}{MineStudio} \citep{MineStudio}. 

\section{Acknoledgements}
This work was supported by the National Science and Technology Major Project \#2022ZD0114902 and the CCF-Baidu Open Fund. We sincerely appreciate their generous support, which enabled us to conduct this research.

\clearpage
\bibliographystyle{abbrvnat}
\bibliography{refs}

\end{document}